\documentclass[10pt,twocolumn,letterpaper]{article}

\usepackage{iccv}
\usepackage{times}
\usepackage{epsfig}
\usepackage{graphicx}
\usepackage{amsmath}
\usepackage{amssymb}

\usepackage{algorithm}
\usepackage{algpseudocode}
\usepackage{multirow}
\usepackage{epstopdf}
\usepackage{authblk}

\usepackage{mmstyle}

\newcommand{\SSN}{structured segment network}

\usepackage[pagebackref=true,breaklinks=true,letterpaper=true,colorlinks,bookmarks=false]{hyperref}

\iccvfinalcopy 

\def\iccvPaperID{1218} 

\ificcvfinal\fi
\begin{document}

\title{Temporal Action Detection with Structured Segment Networks}

\author[1]{Yue Zhao}
\author[1]{Yuanjun Xiong}
\author[2]{Limin Wang}
\author[1]{Zhirong Wu}
\author[1]{Xiaoou Tang}
\author[1]{Dahua Lin}

\affil[1]{Department of Information Engineering, The Chinese University of Hong Kong}
\affil[2]{Computer Vision Laboratory, ETH Zurich, Switzerland}

\maketitle


\begin{abstract}
Detecting actions in untrimmed videos is an important yet
challenging task. 
In this paper, we present the structured segment network (SSN), a novel framework
which models the temporal structure of each action instance
via a structured temporal pyramid.
On top of the pyramid, we further introduce a decomposed discriminative model comprising two classifiers, respectively for classifying actions
and determining completeness. This allows the framework to effectively
distinguish positive proposals from background or incomplete ones,
thus leading to both accurate recognition and localization.
These components are integrated into a unified network that
can be efficiently trained in an end-to-end fashion.
Additionally, a simple yet effective
temporal action proposal scheme, dubbed temporal actionness grouping (TAG) is devised to generate high quality action proposals.
On two challenging benchmarks,
THUMOS’14 and ActivityNet, our method remarkably outperforms previous
state-of-the-art methods, demonstrating superior accuracy and strong adaptivity
in handling actions with various temporal structures.
\footnote{Code available at \url{http://yjxiong.me/others/ssn}}
\end{abstract}




\section{Introduction}
\label{sec:intro}

Temporal action detection has drawn increasing attention from the research community,
owing to its numerous potential applications in surveillance, video analytics, and other areas~\cite{Oneata2013FV,Mettes2015Bofrag,Yeung2016FrameGlimpse,Shou2016SCNN}.
This task is to detect human action instances from untrimmed, and possibly very long videos.
Compared to action recognition, it is substantially more challenging,
as it is expected to output not only the action category, but also the precise starting and ending time points.

\begin{figure}
	\centering
	\includegraphics[height=.80\linewidth]{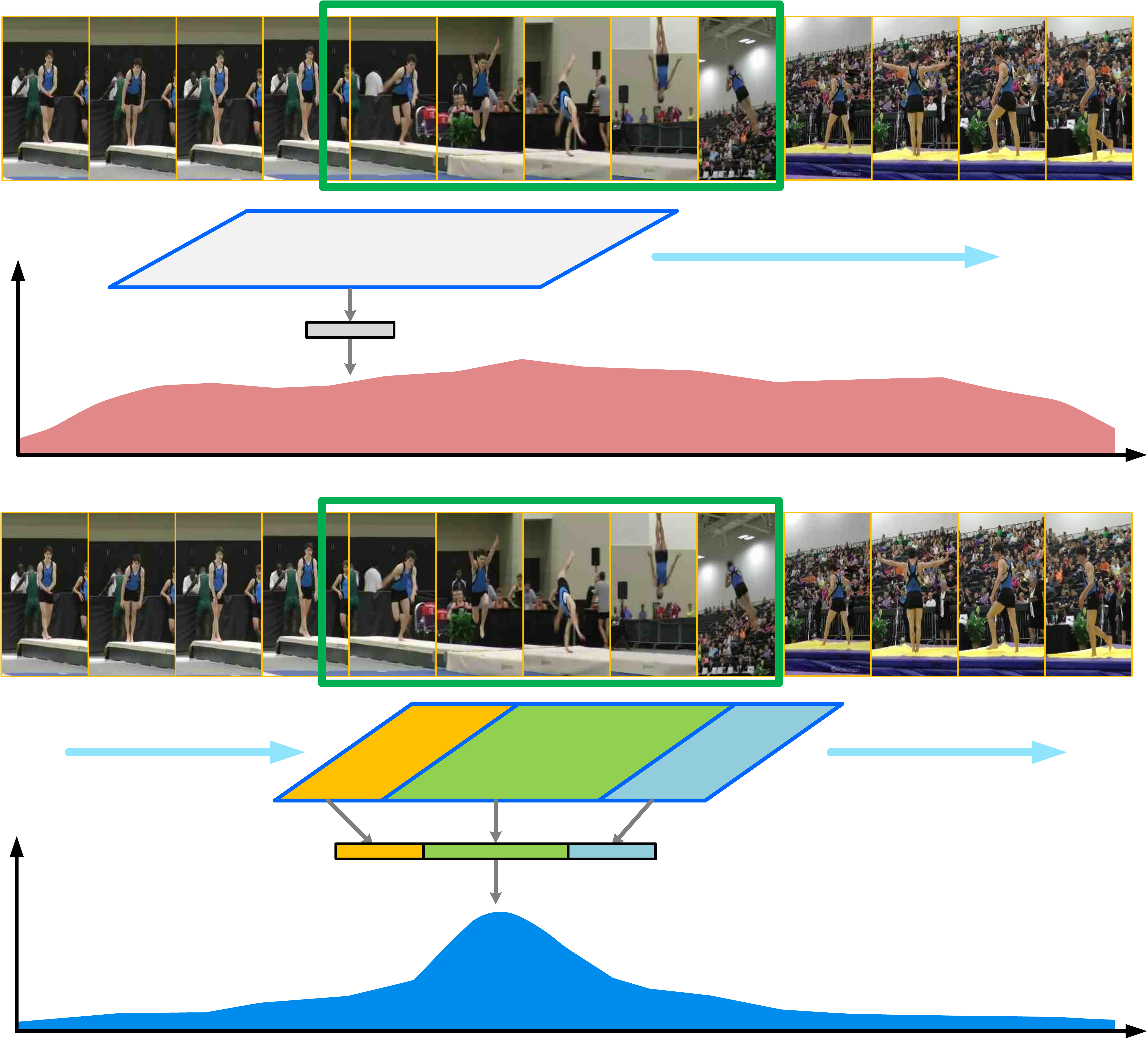}
	\caption{
		\small
		Importance of modeling stage structures in action detection.
		We slide window detectors through a video clip with an action instance of ``Tumbling'' (green box). \textbf{Top}: The detector builds features without any stage structure of the action, \emph{e.g.} average pooling throughout the window. It produces high responses whenever it sees any discriminative snippet related to tumbling, making it hard to localize the instance.
		\textbf{Bottom}: SSN detector utilizes stage structures (starting, course, and ending) via structured temporal pyramid pooling. Its response is only significant when the window is well aligned.}
	\label{fig:teaser_curve}
\end{figure}

Over the past several years, the advances in convolutional neural networks
have led to remarkable progress in video analysis. Notably, the accuracy
of action recognition has been significantly
improved~\cite{Simonyan14TwoStream,Tran15C3D,Fernando15Evolution,WangQT15TDD,Wang2016TSN}.
Yet, the performances of action detection methods remain unsatisfactory~\cite{Yuan2016ScorePyramids,Yeung2016FrameGlimpse,DBLP:journals/corr/SinghC16}.
For existing approaches, one major challenge in precise temporal localization is the large number of incomplete action fragments in the proposed temporal regions. 
Traditional snippet based classifiers rely on discriminative snippets of actions, which would also exist in these incomplete proposals. This makes them very hard to distinguish from valid detections (see Fig.~\ref{fig:teaser_curve}).
We argue that tackling this challenge requires the capability of temporal structure analysis,
or in other words, the ability to identify different stages
\eg~\emph{starting}, \emph{course}, and \emph{ending}, which together decide the \emph{completeness} of an actions instance.

Structural analysis is not new in computer vision.
It has been well studied in various tasks,
\eg~image segmentation~\cite{lafferty2001conditional}, scene understanding~\cite{hoiem2008putting}, and human pose estimation~\cite{andriluka2009pictorial}.
Take the most related object detection for example, in deformable part based models (DPM)~\cite{Felzenszwalb2010DPM},
the modeling of the spatial configurations among parts is crucial.
Even with the strong expressive power of convolutional networks~\cite{Girshick2014RCNN},
explicitly modeling spatial structures, in the form of spatial pyramids~\cite{Lazebnik2006Beyond,He2014SPP}, remains an effective way to achieve improved performance, as demonstrated in a number of
state-of-the-art object detection frameworks, \eg~Fast R-CNN~\cite{Girshick2015FRCNN} and region-based FCN~\cite{Li2016RFCN}.

In the context of video understanding, although temporal structures have played an crucial role in action recognition~\cite{Niebles2010Modeling,Wang2014Latent,Pirsiavash2014SegGrammar,Wang2016TPP}, their modeling in temporal action detection was not as common and successful.
Snippet based methods~\cite{Mettes2015Bofrag,DBLP:journals/corr/SinghC16}
often process individual snippets independently without considering the temporal
structures among them.
Later works attempt to incorporate temporal structures, but are often limited to
analyzing short clips.
S-CNN~\cite{Shou2016SCNN} models the temporal structures via the 3D convolution,
but its capability
is restricted by the underlying architecture~\cite{Tran15C3D}, which is designed to
accommodate only $16$ frames.
The methods based on recurrent networks~\cite{DonahueJ2015LRCN,Montes_2016_NIPSWS}
rely on dense snippet sampling and thus are confronted with serious computational
challenges when modeling long-term structures.
Overall, existing works are limited in two key aspects.
First, the tremendous amount of visual data in videos restricts their capability of modeling long-term dependencies in an end-to-end manner.
Also, they neither provide \emph{explicit} modeling of different
stages in an activity (\eg~\emph{starting} and \emph{ending}) nor offer a mechanism
to assess the \emph{completeness}, which, as mentioned, is crucial for accurate action
detection.

In this work, we aim to move beyond these limitations and develop an effective technique for temporal action detection.
Specifically, we adopt the proven paradigm of ``proposal+classification'', but take
a significant step forward by utilizing explicit structural modeling in the temporal dimension.
In our model, each complete activity instance is considered as a composition of
three major stages, namely \emph{starting}, \emph{course}, and \emph{ending}.
We introduce structured temporal pyramid pooling to
produce a global representation of the entire proposal.
Then we introduce a decomposed discriminative model to jointly classify action categories and determine \emph{completeness} of the proposals, which work collectively to output only complete action instances. 
These components are integrated
into a unified network, called \emph{structured segment network} (SSN).
We adopt the sparse snippet sampling strategy~\cite{Wang2016TSN}, which overcomes
the computational issue for long-term modeling and
enables efficient end-to-end training of SSN.
Additionally, we propose to use multi-scale grouping upon the temporal actionness signal to generate action proposals, achieving higher temporal recall with less proposals to further boost the detection performance.

The proposed SSN framework excels in the following aspects:
1) It provides an effective mechanism to model the temporal structures of activities,
and thus the capability of discriminating between complete and incomplete proposals.
2) It can be efficiently learned in an end-to-end fashion
({$5$ to $15$} hours over a large video dataset, \eg~ActivityNet),
and once trained, can perform fast inference of temporal structures.
3) The method achieves superior detection performance on standard benchmark datasets, establishing new state-of-the-art for temporal action detection.


\section{Related Work}
\label{related}

\noindent \textbf{Action Recognition.}
Action recognition has been extensively studied in the past few years~\cite{Laptev05STIP,WangS13IDT,Simonyan14TwoStream,Tran15C3D,WangQT15TDD,Wang2016TSN,ZhangWW0W16}.
Earlier methods are mostly based on hand-crafted visual features~\cite{Laptev05STIP,WangS13IDT}.
In the past several years, the wide adoption of convolutional networks (CNNs)
has resulted in remarkable performance gain.
CNNs are first introduced to this task in~\cite{KarpathyCVPR14Sports1M}.
Later, two-stream architectures~\cite{Simonyan14TwoStream} and 3D-CNN~\cite{Tran15C3D}
are proposed to incorporate both appearance and motion features.
These methods are primarily frame-based and snippet-based, with simple schemes to aggregate results.
There are also efforts that explore long-range temporal structures via temporal pooling or RNNs~\cite{WangQT15TDD,Ng15BeyondSnippet,DonahueJ2015LRCN}.
However, most methods assume well-trimmed videos, where
the action of interest lasts for nearly the entire duration.
Hence, they don't need to consider the issue of localizing the action instances.

\begin{figure*}[t]
	\centering
	\includegraphics[width=0.96\linewidth]{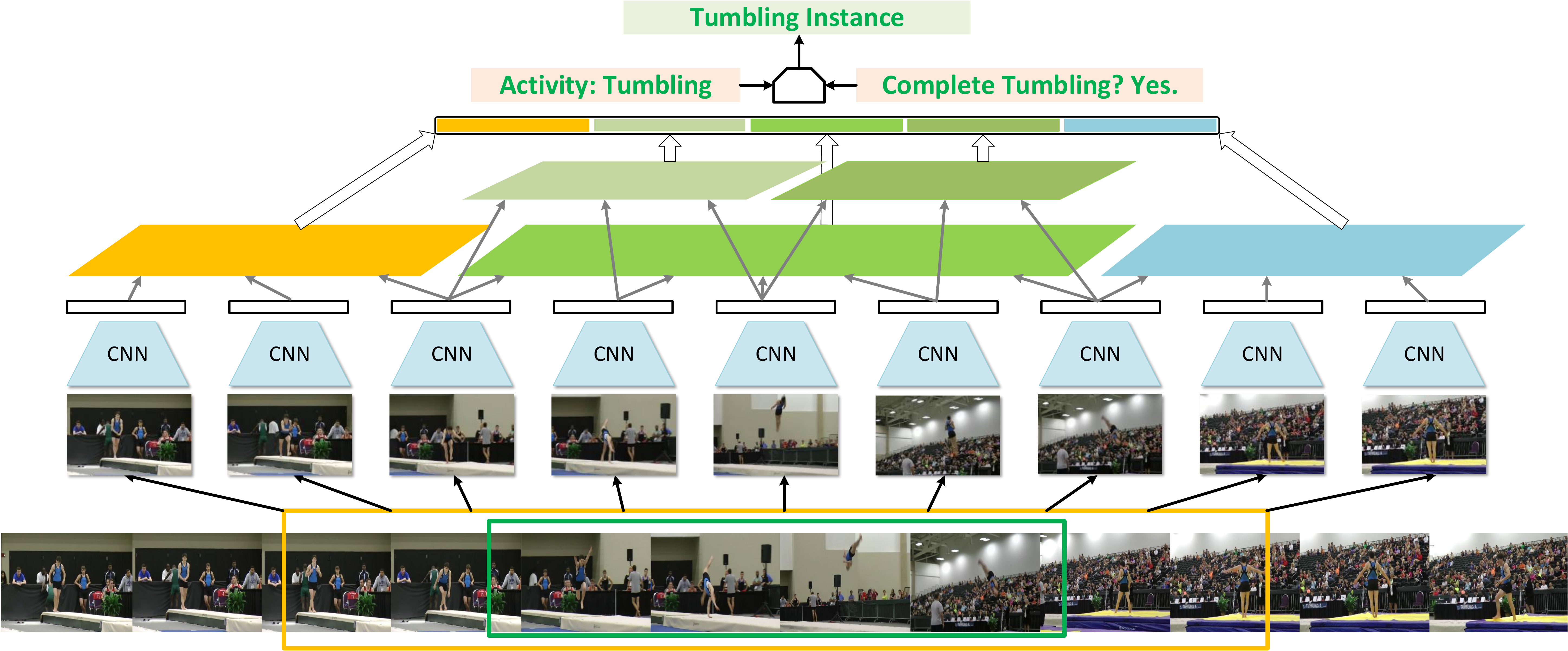}
	\caption{\small
		An overview of the structured segment network framework. On a video from ActivityNet~\cite{caba2015activitynet} there is a candidate region (green box). We first build the augmented proposal (yellow box) by extending it.
		The augmented proposal is divided into starting (orange), course (green), and ending (blue) stages. An additional level of pyramid with two sub-parts is constructed on the course stage. Features from CNNs are pooled within these five parts and concatenated to form the global region representations.
		The activity classifier and the completeness classifier operate on the the region representations to produce activity probability and class conditional completeness probability.
		The final probability of the proposal being positive instance is decided by the joint probability from these two classifiers.
		During training, we sparsely sample $ L=9 $ snippets from evenly divided segments to approximate the dense temporal pyramid pooling.}
	\label{fig:overview}
\end{figure*}

\noindent \textbf{Object Detection.}
Our action detection framework is closely related to object detection frameworks~\cite{Felzenszwalb2010DPM,Girshick2014RCNN,Ren2015FasterRCNN} in spatial images, where detection is performed by classifying object proposals into foreground classes and a background class.
Traditional object proposal methods rely on dense sliding windows~\cite{Felzenszwalb2010DPM} and bottom-up methods that exploit low-level boundary cues~\cite{Van2011SS,Dollar2014Edgebox}.
Recent proposal methods based on deep neural networks show better average recall while requiring less candidates~\cite{Ren2015FasterRCNN}.
Deep models also introduce great modeling capacity for capturing object appearances.
With strong visual features, spatial structural modeling~\cite{Lazebnik2006Beyond} remains a key component for detection.
In particular, the RoI pooling~\cite{Girshick2015FRCNN} is introduced to model the spatial configuration of object with minimal extra cost.
The idea is further reflected in R-FCN~\cite{Li2016RFCN} where the spatial configuration is handled with the position sensitive pooling.

\noindent \textbf{Temporal Action Detection.}
Previous works on activity detection mainly use sliding windows as candidates
and focus on designing hand-crafted feature representations for classification~\cite{Gaidon2013Actom,Tang2013RightFeature,Oneata2013FV,Mettes2015Bofrag,Yuan2016ScorePyramids,JainCVPR14Tubelet}.
Recent works incorporate deep networks into the detection frameworks and obtain improved performance~\cite{Yeung2016FrameGlimpse,Shou2016SCNN,de2016online}.
S-CNN~\cite{Shou2016SCNN} proposes a multi-stage CNN which boosts accuracy via a localization network.
However, S-CNN relies on C3D~\cite{Tran15C3D} as the feature extractor,
which is initially designed for snippet-wise action classification.
Extending it to detection with possibly long action proposals needs
enforcing an undesired large temporal kernel stride.
Another work~\cite{Yeung2016FrameGlimpse} uses Recurrent Neural Network (RNN) to learn a glimpse policy for predicting the starting and ending points of an action.
Such sequential prediction is often time-consuming for processing long videos and
it does not support joint training of the underlying feature extraction CNN.
Our method differs from these approaches in that
it explicitly models the action structure via structural temporal pyramid pooling.
By using sparse sampling, we further enable efficient end-to-end training.
Note there are also works on spatial-temporal detection~\cite{Gkioxari2015Tubes,Weinzaepfel2015Track,mettes2016spot,Wang2016Actionness,Peng2016ActionFRCNN} and
temporal video segmentation~\cite{Hoai2011Joint},
which are beyond the scope of this paper.


\section{Structured Segment Network}
\label{sec:overview}

The proposed \SSN~framework, as shown in Figure~\ref{fig:overview},
takes as input a video and a set of temporal action proposals.
It outputs a set of predicted \emph{activity instances}
each associated with a category label and a temporal range (delimited
by a starting point and an ending point).
From the input to the output, it takes three key steps.
First, the framework relies on a proposal method to produce a set of \emph{temporal proposals}
of varying durations, where each proposal comes with a starting and an ending time.
The proposal methods will be discussed in detail in Section~\ref{sec:proposal}.
Our framework considers each proposal as a composition of
three consecutive stages, \emph{starting}, \emph{course}, and \emph{ending},
which respectively capture how the action starts, proceeds, and ends.
Thus upon each proposal, structured temporal pyramid pooling (STPP) are performed by 1) splitting the proposal into the three stages; 2) building temporal pyramidal representation for each stage; 3) building global representation for the whole proposal by concatenating stage-level representations.
Finally, two classifiers respectively for recognizing the activity category
and assessing the completeness will be applied on the representation obtained by STPP and their predictions
will be combined, resulting in a subset of \emph{complete} instances tagged with category labels.
Other proposals, which are considered as either \emph{belonging to background} or
\emph{incomplete}, will be filtered out.
All the components outlined above are integrated into a unified network,
which will be trained in an end-to-end way.
For training, we adopt the sparse snippet sampling strategy~\cite{Wang2016TSN}
to approximate the temporal pyramid on dense samples.
By exploiting the redundancy among video snippets, this strategy can substantially
reduce the computational cost, thus allowing the crucial modeling of long-term temporal
structures.

\vspace{-5pt}
\subsection{Three-Stage Structures}

At the input level, a video can be represented as a sequence of $T$ \emph{snippets},
denoted as $(S_t)_{t=1}^T$.
Here, one snippet contains several consecutive frames, which, as a whole, is
characterized by a combination of RGB images and an optical flow stack~\cite{Simonyan14TwoStream}.
Consider a given set of $N$ proposals $P=\{p_i=[s_i, e_i]\}_{i=1}^N$.
Each proposal $p_i$ is composed of a starting time $s_i$ and an ending time $e_i$.
The duration of $ p_i $ is thus $ d_i = e_i - s_i $.
To allow structural analysis and particularly to determine whether a proposal
captures a \emph{complete} instance, we need to put it in a context.
Hence, we augment each proposal $p_i$  into
$p'_i=[s'_i, e'_i]$ with
where $s'_i = s_i - d_i / 2$ and $e'_i = e_i + d_i/2$.
In other words, the augmented proposal $p'_i$ doubles the span of $p_i$ by
extending beyond the starting and ending points, respectively by $d_i / 2$.
If a proposal accurately aligns well with a groundtruth instance, the augmented proposal
will capture not only the inherent process of the activity, but also how it starts and ends.
Following the three-stage notion, we divide the augmented proposal $p'_i$ into three
consecutive intervals:
$p_i^s = [s'_i, s_i]$, $p_i^c = [s_i, e_i]$, and $p_i^e = [e_i, e'_i]$, which
are respectively corresponding to the \emph{starting}, \emph{course}, and \emph{ending} stages.

\subsection{Structured Temporal Pyramid Pooling}
\label{sec:stpp}

As mentioned, the \SSN~framework derives a global representation for
each proposal via temporal pyramid pooling.
This design is inspired by the success of spatial pyramid
pooling~\cite{Lazebnik2006Beyond,He2014SPP} in object recognition
and scene classification.
Specifically, given an augmented proposal $p'_i$ divided into three
stages $p_i^s$, $p_i^c$, and $p_i^e$, we first compute the stage-wise
feature vectors $\vf_i^s$, $\vf_i^c$, and $\vf_i^e$ respectively
via temporal pyramid pooling, and then concatenate them into
a global representation.

Specifically, a stage with interval $[s, e]$ would cover a series
of snippets, denoted as $\{S_t | s \le t \le e\}$. For each snippet,
we can obtain a feature vector $\vv_t$.
Note that we can use any feature extractor here. In this work, we
adopt the effective two-stream feature representation first proposed in~\cite{Simonyan14TwoStream}.
Based on these features, we construct a $L$-level temporal pyramid where each level evenly divides the interval into $B_l$ parts.
For the $i$-th part of the $l$-th level, whose interval is $[s_{li}, e_{li}]$,
we can derive a pooled feature as
\begin{equation}
    \vu_i^{(l)} = \frac{1}{|e_{li} - s_{li} + 1|}
    \sum_{t=s_{li}}^{e_{li}} \vv_t.
\end{equation}
Then the overall representation of this stage can be obtained by
concatenating the pooled features across all parts at all levels
as
$
	\vf_i^c = (\vu_i^{(l)} | l=1,\ldots,L,~i=1,~\ldots,B_l)
$.

We treat the three stages differently.
Generally, we observed that the \emph{course} stage, which reflects
the activity process itself, usually contains richer structure
\eg~this process itself may contain sub-stages.
Hence, we use a two-level pyramid, \ie~$L = 2, B_1= 1$, and $ B_2=2$, for the \emph{course}
stage, while using simpler one-level pyramids (which essentially
reduce to standard average pooling) for \emph{starting} and \emph{ending}
pyramids.
We found empirically that this setting strikes a good balance
between expressive power and complexity.
Finally, the stage-wise features are combined via concatenation.
Overall, this construction explicitly leverages the structure
of an activity instance and its surrounding context, and thus
we call it \emph{structured temporal pyramid pooling} (STPP).

\subsection{Activity and Completeness Classifiers}

On top of the structured features described above,
we introduce two types of classifiers, an \emph{activity classifier}
and a set of \emph{completeness classifiers}.
Specifically,
the \emph{activity classifier} $A$ classifies input proposals
into $K + 1$ classes, \ie~$K$ activity classes (with labels $1, \ldots, K$)
and an additional \emph{``background''} class (with label $0$).
This classifier restricts its scope to the \emph{course} stage,
making predictions based on the corresponding feature $\vf_i^c$.
The \emph{completeness classifiers} $\{C_k\}_{k=1}^K$ are a set of
binary classifiers, each for one activity class.
Particularly, $C_k$ predicts whether a proposal captures a \emph{complete}
activity instance of class $k$, based on the global representation $\{\vf_i^s, \vf_i^c, \vf_i^e\}$ induced by STPP. In this way, the \emph{completeness}
is determined not only on the proposal itself but also on its
surrounding context.

Both types of classifiers are implemented as linear classifiers
on top of high-level features. Given a proposal $p_i$,
the activity classifier will produce
a vector of normalized responses via a softmax layer.
From a probabilistic view, it can be considered as a conditional
distribution $P(c_i | p_i)$, where $c_i$ is the class label.
For each activity class $k$, the corresponding completeness
classifier $C_k$ will yield a probability value, which can
be understood as the conditional probability $P(b_i | c_i, p_i)$,
where $b_i$ indicates whether $p_i$ is \emph{complete}.
Both outputs together form a joint distribution. When $c_i \ge 1$,
$
    P(c_i, b_i | p_i) =  P(c_i | p_i) \cdot P(b_i | c_i, p_i).
$
Hence, we can define a \emph{unified classification loss}
jointly on both types of classifiers. With a proposal $p_i$
and its label $c_i$:
\begin{equation}
    \cL_{cls}(c_i, b_i; p_i) =
    - \log P(c_i | p_i)
    - 1_{(c_i \ge 1)} \log P(b_i | c_i, p_i).
\end{equation}
Here, the \emph{completeness} term $P(b_i | c_i, p_i)$ is only used when
$c_i \ge 1$, \ie~the proposal $p_i$ is not considered as part of the background.
Note that these classifiers together with STPP
are integrated into a single network that is trained in an \emph{end-to-end} way.

During training, we collect three types of proposal samples:
(1) \emph{positive proposals}, \ie~those overlap
with the closest groundtruth instances with at least $0.7$ IoU;
(2) \emph{background proposals}, \ie~those that do not overlap
with any groundtruth instances; and
(3) \emph{incomplete proposals},
\ie~those that satisfy the following criteria: $80\%$ of its own span is
contained in a groundtruth instance, while its IoU with that instance is
below $0.3$ (in other words, it just covers a small part of the instance).
For these proposal types, we respectively have
$(c_i > 0, b_i = 1)$, $c_i = 0$, and $(c_i > 0, b_i = 0)$.
Each mini-batch is ensured to contain all three types of proposals.

\subsection{Location Regression and Multi-Task Loss}

With the structured information encoded in the global features,
we can not only make categorical predictions, but also refine the
proposal's temporal interval itself by \emph{location regression}.
We devise a set of location regressors $\{R_k\}_{k=1}^K$,
each for an activity class.
We follow the design in RCNN~\cite{Girshick2014RCNN}, but adapting
it for 1D temporal regions.
Particularly, for a \emph{positive proposal} $p_i$, we regress
the relative changes of both the interval center $\mu_i$ and the span $\phi_i$ (in log-scale),
using the closest groundtruth instance as the target.
With both the classifiers and location regressors,
we define a multi-task loss over an training sample $p_i$, as:
\begin{equation}
    \cL_{cls}(c_i, b_i; p_i) +
    \lambda \cdot 1_{(c_i \ge 1 \ \& \ b_i = 1)}
    \cL_{reg}(\mu_i, \phi_i; p_i).
\end{equation}
Here, $\cL_{reg}$ uses the smooth $L_1$ loss function~\cite{Girshick2015FRCNN}.


\section{Efficient Training and Inference with SSN}

The huge amount of frames poses a serious challenge in computational cost
to video analysis. Our \SSN~also faces this challenge.
This section presents two techniques which we use to reduce
the cost and enable end-to-end training.

\vspace{-12pt}
\paragraph{Training with sparse sampling.}
The structured temporal pyramid, in its original form, rely on
densely sampled snippets. This would lead to excessive computational cost
and memory demand in end-to-end training over long proposals -- in practice,
proposals that span over hundreds of frames are not uncommon.
However, dense sampling is generally unnecessary in our framework.
Particularly, the \emph{pooling} operation is essentially to collect
feature statistics over a certain region. Such statistics can be well
approximated via a subset of snippets, due to the high redundancy
among them.

Motivated by this, we devise a \emph{sparse snippet sampling scheme}.
Specifically, given a augmented proposal $p'_i$,
we evenly divide it into $L = 9$ segments,
randomly sampling only one snippet from each segment.
Structured temporal pyramid pooling is performed for each pooling region on its corresponding segments.
This scheme is inspired by the segmental architecture in~\cite{Wang2016TSN},
but differs in that it operates within STPP instead of a global average pooling.
In this way, we fix the number of features needed to be computed regardless
of how long the proposal is, thus effectively reducing the computational cost,
especially for modeling long-term structures.
More importantly, this enables end-to-end training of the entire framework
over a large number of long proposals.

\vspace{-12pt}
\paragraph{Inference with reordered computation.}
In testing, we sample video snippets with a fixed interval of $6$ frames,
and construct the temporal pyramid thereon.
The original formulation of temporal pyramid first computes pooled features
and then applies the classifiers and regressors on top which is not efficient.
Actually, for each video, hundreds of proposals will be generated,
and these proposals can significantly overlap with each
other -- therefore, a considerable portion of the snippets and
the features derived thereon are shared among proposals.

To exploit this redundancy in the computation,
we adopt the idea introduced in position sensitive pooling~\cite{Li2016RFCN}
to improve testing efficiency.
Note that our classifiers and regressors are both linear.
So the key step in classification or regression is to multiply
a weight matrix $\mW$ with the global feature vector $\vf$.
Recall that $\vf$ itself is a concatenation of multiple features,
each pooled over a certain interval. Hence the computation
can be written as $\mW \vf = \sum_{j} \mW_j \vf_j$,
where $j$ indexes different regions along the pyramid.
Here, $\vf_j$ is obtained by \emph{average pooling} over all snippet-wise features
within the region $r_j$. Thus, we have
\begin{equation} \label{eq:reorder}
    \mW_j \vf_j
    = \mW_j \cdot \Ebb_{t \sim r_j}\left[ \vv_t \right]
    = \Ebb_{t \sim r_j} \left[ \mW_j \vv_t \right].
\end{equation}
$\Ebb_{t \sim r_j}$ denotes the average pooling over $r_j$,
which is a linear operation and therefore can be exchanged with
the matrix multiplication.
Eq~\eqref{eq:reorder} suggests that
the linear responses \wrt~the classifiers/regressors can be computed
\emph{before} pooling.
In this way, the heavy matrix multiplication can be done in the CNN for
each video over all snippets, and for each proposal, we only
have to pool over the network outputs.
This technique can reduce the processing time after extracting network outputs from
around 10 seconds to less than 0.5 second per video on average.


\section{Temporal Region Proposals}
\label{sec:proposal}

\begin{figure}[t]
	\centering
	\includegraphics[width=.9\linewidth]{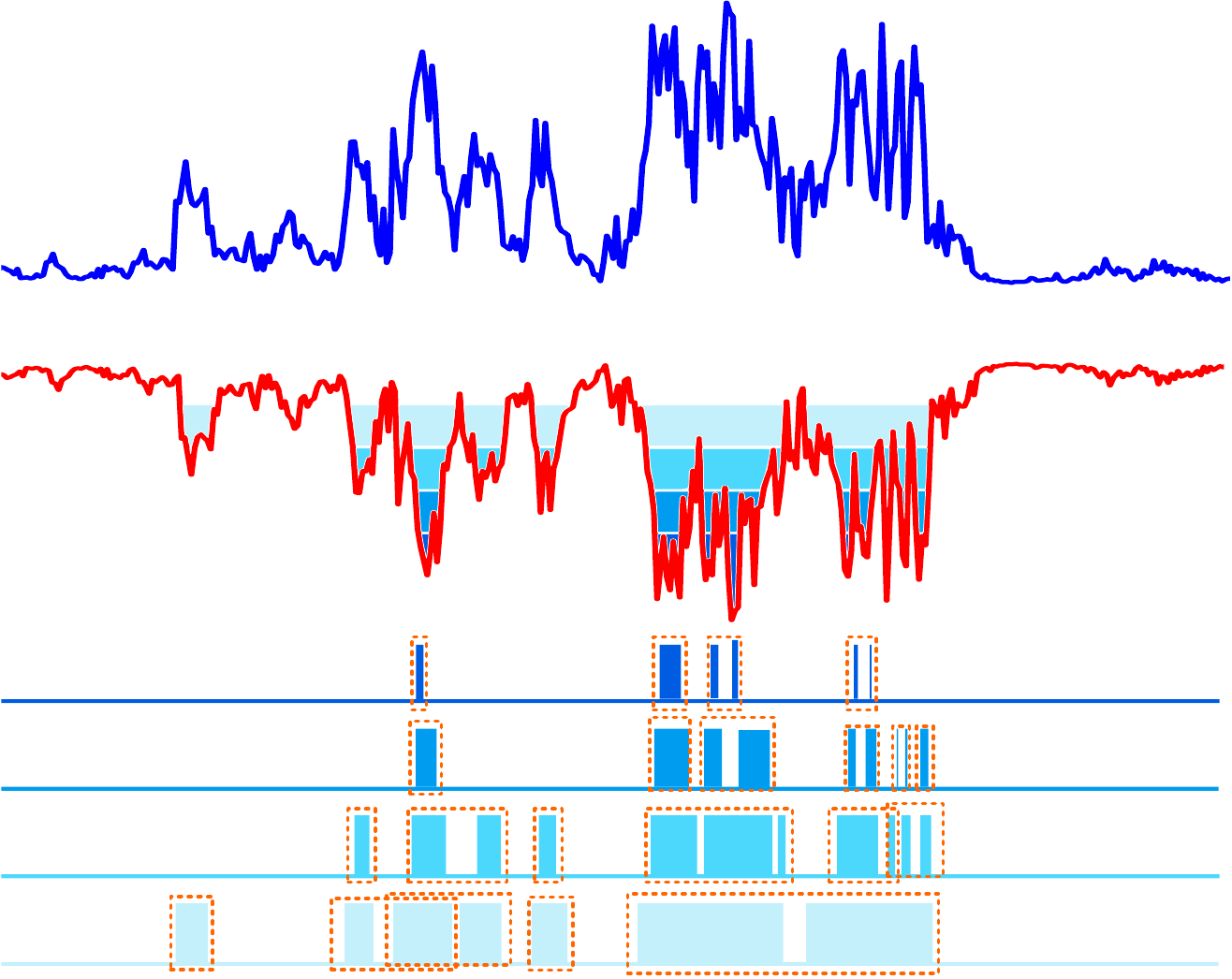}
	\caption{\small
		Visualization of the temporal actionness grouping process for proposal generation.
		\textbf{Top}: Actionness probabilities as a 1D signal sequence.
		\textbf{Middle}: The complement signal. We flood it with different levels $\gamma$.
		\textbf{Bottom}: Regions obtained by different flooding levels.
		By merging the regions according to the grouping criterion, we
		get the final set of proposals (in orange color).
	}
	\label{fig:tag_proposal}
	\vspace{-10pt}
\end{figure}

In general, SSN accepts arbitrary proposals,
\eg~sliding windows~\cite{Shou2016SCNN,Yuan2016ScorePyramids}.
Yet, an effective proposal method can produce more accurate proposals,
and thus allowing a small number of proposals to reach a certain level
of performance. In this work, we devise an effective
proposal method called \emph{temporal actionness grouping (TAG)}.

This method uses an actionness classifier
to evaluate the binary \emph{actionness probabilities} for individual snippets.
The use of binary actionness for proposals is first introduced in spatial action detection by~\cite{Wang2016Actionness}.
Here we utilize it for temporal action detection.

Our basic idea is to find those continuous temporal regions with mostly
high actionness snippets to serve as proposals.
To this end, we repurpose a classic watershed algorithm~\cite{Roerdink2000Watershed},
applying it to the 1D signal formed by a sequence of \emph{complemented}
actionness values, as shown in Figure~\ref{fig:tag_proposal}.
Imagine the signal as 1D terrain with heights and basins. This algorithm floods water on this terrain with different \emph{``water level''}
$(\gamma)$, resulting in a set of \emph{``basins''} covered by water, denoted by $G(\gamma)$.
Intuitively, each ``basin'' corresponds to a temporal region with high actionness.
The ridges above water then form the blank areas between basins, as illustrated in  Fig.~\ref{fig:tag_proposal}.

Given a set of basins $G(\gamma)$, we devise a grouping scheme
similar to~\cite{PT2015MCG}, which tries to connect small basins into
proposal regions. The scheme works as follows: it begins with a
seed basin, and consecutively absorbs the basins that follow,
until the fraction of the basin durations over the total duration
(\ie~from the beginning of the first basin to the ending of the last)
drops below a certain threshold $\tau$. The absorbed basins and the blank spaces between them are then grouped to form a single proposal. 
We treat each basin as seed and perform the grouping procedure to obtain a set of proposals denoted by $G'(\tau, \gamma)$.
Note that we do not choose a specific combination of $\tau$ and $\gamma$.
Instead we uniformly sample $\tau$ and $\gamma$ from  $\in (0, 1)$ with an even step of $ 0.05 $.
The combination of these two thresholds leads to multiple sets of regions.
We then take the \emph{union} of them.
Finally, we apply non-maximal suppression to the union with IoU threshold $0.95$,
to filter out highly overlapped proposals. The retained proposals will be fed to
the SSN framework.


\section{Experimental Results}
\label{sec:experiment}

We conducted experiments to test the proposed framework
on two large-scale action detection benchmark datasets:
\emph{ActivityNet}~\cite{caba2015activitynet} and
\emph{THUMOS14}~\cite{Jiang2014THUMOS14}.
In this section we first introduce these datasets and other experimental settings and then investigate the impact of different components
via a set of ablation studies.
Finally we compare the performance of SSN with other state-of-the-art approaches.

\subsection{Experimental Settings}

\paragraph{Datasets.}
\textbf{ActivityNet}~\cite{caba2015activitynet} has two versions, \emph{v1.2} and \emph{v1.3}.
The former contains $9682$ videos in $100$ classes, while the latter, which is a superset of v1.2 and
was used in the ActivityNet Challenge 2016, contains $19994$ videos in $200$ classes.
In each version, the dataset is divided into three disjoint subsets,
training, validation, and testing, by $2$:$1$:$1$.
\textbf{THUMOS14}~\cite{Jiang2014THUMOS14} has $1010$ videos for validation and $1574$ videos for testing.
This dataset does not provide the training set by itself.
Instead, the UCF101~\cite{Soomro2012Ucf101}, a trimmed video dataset is appointed as the official training set.
Following the standard practice, we train out models on the validation set and evaluate them
on the testing set.
On these two sets, $ 220 $ and $ 212 $ videos have temporal annotations in $20$ classes, respectively. $ 2 $ falsely annotated videos (``270'',``1496'') in the test set are excluded in evaluation.
In our experiments,
we compare with our method with the states of the art on both
\emph{THUMOS14} and \emph{ActivityNet v1.3},
and perform ablation studies on \emph{ActivityNet v1.2}.

\vspace{-12pt}
\paragraph{Implementation Details.}
We train the \SSN~in an end-to-end manner,
with raw video frames and action proposals as the input.
Two-stream CNNs~\cite{Simonyan14TwoStream} are used for feature extraction.
We also use the spatial and temporal streams to harness both the appearance and motion features.
The binary actionness classifiers underlying the TAG proposals are trained with~\cite{Wang2016TSN} on the training subset of each dataset.
We use SGD to learn CNN parameters in our framework, with batch size $ 128 $ and momentum $0.9$.
We initialize the CNNs with pre-trained models from ImageNet~\cite{Deng2009ImageNet}.
The initial learning rates are set to $ 0.001 $ for RGB networks and $ 0.005 $ for optical flow networks.
In each minibatch, we keep the ratio of three types of proposals, namely
\emph{positive}, \emph{background}, and \emph{incomplete}, to be $1$:$1$:$6$.
For the completeness classifiers, only the samples with loss values ranked in the first $ 1/6 $ of a minibatch are used for calculating gradients, which resembles online hard negative mining~\cite{Shrivastava2016OHNM}.
On both versions of ActivityNet, the RGB and optical flow branches of the two-stream CNN
are respectively trained for $9.5K$ and $20K$ iterations,
with learning rates scaled down by $0.1$ after every $4K$ and $8K$ iterations, respectively.
On THUMOS14, these two branches are respectively trained for $1K$ and $6K$ iterations,
with learning rates scaled down by $0.1$ per $400$ and $2500$ iterations.

\vspace{-12pt}
\paragraph{Evaluation Metrics.}
As both datasets originate from contests,
each dataset has its own convention of reporting performance metrics.
We follow their conventions, reporting mean average precision (mAP) at different IoU thresholds.
On both versions of ActivityNet, the IoU thresholds are $ \{0.5, 0.75, 0.95\} $.
The average of mAP values with IoU thresholds $[0.5$:$0.05$:$0.95]$ is used to compare the performance between different methods.
On THUMOS14, the IoU thresholds are $\{0.1, 0.2, 0.3, 0.4, 0.5\}$.
The mAP at $0.5$ IoU is used for comparing results from different methods.

\subsection{Ablation Studies}

\begin{figure}[t]
	\begin{center}
		\includegraphics[width=0.7\linewidth]{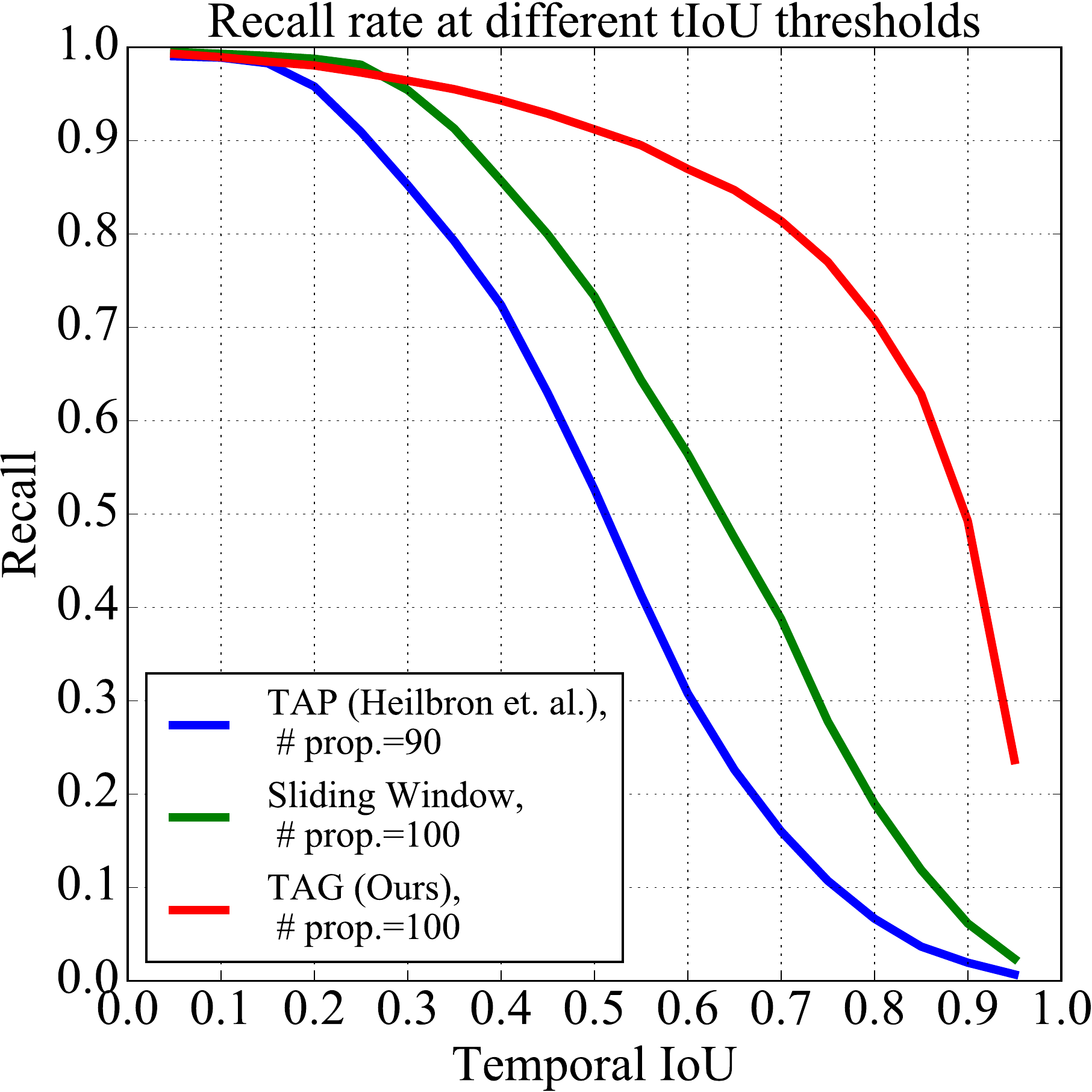}
	\end{center}
	\caption{\small
		Recall rate at different tIoU thresholds on ActivityNet v1.2.
		High recall rates at high IoU  thresholds ($>0.7$) indicate better proposal quality.}
	\label{fig:recall_iou}
\vspace{-5pt}
\end{figure}

\begin{table}[t]
\begin{center}
		\begin{tabular}{c|c|c|c|c}
			\hline
			\multirow{2}{*}{Proposal Method}& \multicolumn{2}{|c|}{\textbf{THUMOS14}}  & \multicolumn{2}{|c}{\textbf{ActivityNet v1.2}} \\
			\cline{2-5}
			& \# Prop. & AR & \# Prop. & AR \\
			\hline
			Sliding Windows&  204  & 21.2 & 100 & 34.8\\
			\hline
			SCNN-prop~\cite{Shou2016SCNN} & 200 & 20.0 & - & - \\
			\hline
			TAP~\cite{caba2016cvpr}& 200  & 23.0  & 90  & 14.9 \\
			\hline
			DAP~\cite{Escorcia2016DAP}& 200 & 37.0 &100 & 12.1  \\ 
			\hline\hline
			TAG & 200 & {\bf 48.9} & 100 & {\bf 71.7} \\
			\hline
		\end{tabular}
\end{center}
\caption{\small
	Comparison between different temporal action proposal methods with same number of proposals.
	``AR'' refers to the average recall rates.
	``-'' indicates the result is not available.}
\label{table:proposals}
\vspace{-12pt}
\end{table}

\paragraph{Temporal Action Proposal.}

We compare the performance of different action proposal schemes in three aspects, \emph{i.e.} recall, quality, and detection performance.
Particularly, we compare our TAG scheme with common sliding windows as well as other state-of-the-art proposal methods, including
SCNN-prop, a proposal networks presented in~\cite{Shou2016SCNN},
TAP~\cite{caba2016cvpr}, DAP~\cite{Escorcia2016DAP}.
For the sliding window scheme, we use $20$ exponential scales starting from $0.3$ second long
and step sizes of $0.4$ times of window lengths.

We first evaluate the average recall rates, which are summarized in Table~\ref{table:proposals}.
We can see that TAG proposal have higher recall rates with the same number of proposals.
Then we investigate the quality of its proposals.
We plot the recall rates from different proposal methods at different IoU thresholds in Fig.~\ref{fig:recall_iou}.
We can see TAG retains relatively high recall at high IoU thresholds,
demonstrating that the proposals from TAG are generally more accurate.
In experiments we also tried applying the actionness classifier trained on ActivityNet v1.2 directly on THUMOS14.
We can still achieve a reasonable average recall of $ 39.6\% $, while the one trained on THUMOS14 achieves $ 48.9\% $ in Table~\ref{table:proposals}.
Finally, we evaluate the proposal methods in the context of action detection.
The detection mAP values using sliding window proposals and TAG proposals are shown in Table~\ref{table:component}.
The results confirm that, in most cases, the improved proposals can result in improved detection performance.

\vspace{-12pt}
\paragraph{Structured Temporal Pyramid Pooling.}~\label{sec:stpp_tpp}
Here we study the influence of different pooling strategies in STPP.
We denote one pooling configuration as $ (B_1,\ldots,B_K)- A $, where $ K $ refers to the number of pyramid levels for the course stage and $ B_1,\ldots,B_K $ the number of regions in each level. $A=1$ indicates we use augmented proposal and model the starting and ending stage, while $A=0$ indicates we only use the original proposal (without augmentation).
Additionally we compare two within-region pooling methods: average and max pooling.
The results are summarized in Table~\ref{table:pooling}.
Note that these configurations are evaluated in the stage-wise training scenario.
We observe that cases where $ A=0 $ have inferior performance, showing that the introduction of the stage structure is very important for accurate detection.
Also, increasing the depth of the pyramids for the course stage can give slight performance gain.
Based on these results, we fix the configuration to $(1,2)-1$ in later experiments.

\begin{table}[t]
\begin{center}
	\begin{tabular}{c|c|c|c|c}
		\hline
		Average mAP (\%) & (1)-0 & (1,2)-0 &(1)-1 &(1,2)-1 \\ \hline
		Max Pool         & 13.1   & 13.5     & 18.3 &18.4     \\ \hline
		Average Pool     & 4.48    & 4.34      & 24.3 &24.6    \\ \hline
	\end{tabular}
\end{center}
	\caption{\small
		Comparison between different temporal pooling settings.
		The setting  (1,2)-1 is used in the SSN framework. Please refer to Sec.~\ref{sec:stpp_tpp} for the definition of these settings.}
	\label{table:pooling}
	\vspace{-12pt}
\end{table}

\vspace{-12pt}
\paragraph{Classifier Design.}
In this work, we introduced the activity and completeness classifiers which work together to classify the proposal.
We verify the importance of this decomposed design by studying another design
that replaces it with a single set of classifiers, for which
both \emph{background} and \emph{incomplete} samples are uniformly treated as negative. We perform similar negative sample mining for this setting.
The results are summarized in Table~\ref{table:component}.
We observe that using only one classifier to distinguish \emph{positive} samples
from both \emph{background} and \emph{incomplete} would lead to 
worse result even with negative mining, where mAP decreased from $23.7\%$ to $17.9\%$.
We attribute this performance gain to the different natures of the two negative proposal types, which require different classifiers to handle.

\vspace{-12pt}
\paragraph{Location Regression \& Multi-Task Learning.}
Because of the contextual information contained in the starting and ending stages of the global region features, we are able to perform location regression.
We measure the contribution of this step to the detection performance in Table~\ref{table:component}.
From the results we can see that the location regression and multi-task learning, where we train the classifiers and the regressors together in an end-to-end manner, always improve the detection accuracy.

\vspace{-12pt}
\paragraph{Training: Stage-wise v.s. End-to-end.}
While the \SSN~is designed for end-to-end training, it is also possible to first densely extract features and train the classifiers and regressors with SVM and ridge regression, respectively.
We refer to this training scheme as stage-wise training.
We compare the performance of end-to-end training and stage-wise training in Table~\ref{table:component}.
We observe that models from end-to-end training can slightly outperform those learned with stage-wise training under the same settings.
This is remarkable as we are only sparsely sampling snippets in end-to-end training, which also demonstrates the importance of jointly optimizing the classifiers and feature extractors and justifies our framework design. 
Besides, end-to-end training has another major advantage that it does not need to store the extracted features for the training set, which could become quite storage intensive as training data grows.

\begin{table}[t]
\begin{center}
\setlength{\tabcolsep}{4.8pt}¬
\begin{tabular}{c|cccc|cc}
	\hline
	\multicolumn{1}{c|}{} & \multicolumn{4}{c|}{Stage-Wise}            & \multicolumn{2}{c}{End-to-End} \\ \hline
	STPP                  &     & \checkmark & \checkmark & \checkmark & \checkmark     & \checkmark    \\
	Act. + Comp.    &     &            & \checkmark & \checkmark & \checkmark     & \checkmark    \\
	Loc. Reg.   &     &            &            & \checkmark &                & \checkmark    \\ \hline
	SW       & 0.56 &  5.99  & 16.4         &    18.1      & -             &    -         \\ \hline
	TAG                   & 4.82 & 17.9         & 24.6         & 24.9         & 24.8             & 25.9            \\ \hline
\end{tabular}
\end{center}
\caption{\small
	Ablation study on ActivityNet~\cite{caba2015activitynet} v1.2. 
	Overall, end-to-end training is compared against stage wise training.
	We evaluate the performance using both sliding window proposals (``SW'') and TAG proposals (``TAG''), measured by mean average precision (mAP).
	Here, ``STPP'' refers to structure temporal pyramid pooling.
	``Act. + Comp.'' refers to the use of two classifiers design. 
	``Loc. Reg'' denotes the use the location regression.}
\label{table:component}
\vspace{-12pt}
\end{table}

\subsection{Comparison with the State of the Art}

Finally, we compare our method with other state-of-the-art temporal action detection methods on THUMOS14~\cite{Jiang2014THUMOS14} and ActivityNet v1.3~\cite{caba2015activitynet},
and report the performances using the metrics described above.
Note that the average action duration in THUMOS14 and ActivityNet are $4$ and $50$ seconds.
And the average video duration are $233$ and $114$ seconds, respectively.
This reflects the distinct natures of these datasets
in terms of the granularities and temporal structures of the action instances.
Hence, strong adaptivity is required to perform consistently well on both datasets.

\vspace{-12pt}
\paragraph{THUMOS14.}
On THUMOS 14,
We compare with the contest results~\cite{wang2014action,oneata2014lear,Richard2016Language}
and those from recent works, including the methods that use segment-based 3D CNN~\cite{Shou2016SCNN},
score pyramids~\cite{Yuan2016ScorePyramids}, and recurrent reinforcement learning~\cite{Yeung2016FrameGlimpse}.
The results are shown in Table~\ref{table: thumos14}.
In most cases, the proposed method outperforms previous state-of-the-art methods by over $10\%$ in absolute mAP values.

\begin{table}[t]
	\begin{center}
		\begin{tabular}{c|ccccc}
			\hline
			\multicolumn{6}{c}{\textbf{THUMOS14}, \textbf{mAP@$\alpha$}}                 \\ \hline
			Method& 0.1  & 0.2  & 0.3  & 0.4  & 0.5  \\ \hline
			Wang \emph{et. al.}~\cite{wang2014action} & 18.2 & 17.0 & 14.0 & 11.7 & 8.3 \\ \hline
			Oneata \emph{et. al.}~\cite{oneata2014lear} & 36.6 & 33.6 & 27.0 &  20.8 & 14.4 \\ \hline
			Richard \emph{et. al.}~\cite{Richard2016Language} & 39.7 & 35.7 & 30.0 & 23.2 & 15.2 \\ \hline\hline
			S-CNN~\cite{Shou2016SCNN} & 47.7 & 43.5 & 36.3 & 28.7 & 19.0 \\ \hline
			Yeung \emph{et. al.}~\cite{Yeung2016FrameGlimpse} & 48.9 & 44.0 & 36.0 & 26.4 & 17.1 \\ \hline
			Yuan \emph{et. al.}~\cite{Yuan2016ScorePyramids} & 51.4 & 42.6& 33.6&26.1&18.8 \\\hline\hline
			SSN & \textbf{60.3} & \textbf{56.2} & \textbf{50.6} & \textbf{40.8} & \textbf{29.1} \\
			\hline
			SSN* & \textbf{66.0} & \textbf{59.4} & \textbf{51.9} & \textbf{41.0} & \textbf{29.8} \\
			\hline
		\end{tabular}
	\end{center}
	\caption{\small
		Action detection results on THUMOS’14, measured by mAP at different IoU thresholds $\alpha$. The upper half of the table shows challenge results back in 2014. ``SSN*'' indicates metrics calculated in the PASCAL-VOC style used by ActivityNet~\cite{caba2015activitynet}.
	}
	\label{table: thumos14}
	\vspace{-5pt}
\end{table}

\begin{table}[t]
	\begin{center}
		\begin{tabular}{c|ccc|c}
			\hline
			\multicolumn{5}{c}{\textbf{ActivityNet v1.3} (testing),  \textbf{mAP@$\alpha$}}   \\ \hline
			Method & 0.5 & 0.75 & 0.95 &Average\\ \hline
			Wang \emph{et. al.}~\cite{UTS}  & 42.48 & 2.88 & 0.06 & 14.62 \\ \hline
			Singh \emph{et. al.}~\cite{Singh_2016_CVPR} & 28.67 & 17.78 & 2.88 & 17.68 \\ \hline
			Singh \emph{et. al.}~\cite{DBLP:journals/corr/SinghC16}& 36.40 & 11.05 & 0.14 & 17.83 \\ \hline\hline
			SSN & {\bf 43.26} & {\bf 28.70} & {\bf 5.63} &   {\bf 28.28}     \\ \hline
		\end{tabular}
	\end{center}
	\caption{\small
		Action detection results on ActivityNet v1.3, measured by mean average precision (mAP) for different IoU thresholds $\alpha$ and the average mAP of IoU thresholds from $ 0.5 $ to $ 0.95 $.
	}
	\label{table:anet_v1.3}
\vspace{-12pt}
\end{table}

\vspace{-12pt}
\paragraph{ActivityNet.}
The results on the testing set of ActivityNet v1.3 are shown in Table~\ref{table:anet_v1.3}.
For references, we list the performances of highest ranked entries in the ActivityNet 2016 challenge.
We submit our results to the test server of ActivityNet v1.3 and report the detection performance on the testing set.
The proposed framework, using a single model instead of an ensemble, is able to achieve an average mAP of $ 28.28 $ and perform well at high IOU thresholds, \emph{i.e.}, $0.75$ and $0.95$.
This clearly demonstrates the superiority of our method.
Visualization of the detection results can be found in the supplementary materials~\cite{Supplement}.

\vspace{-5pt}
\section{Conclusion}
\label{sec:conclusion}
\vspace{-5pt}
In this paper, we presented a generic framework for temporal action detection,
which combines a structured temporal pyramid with two types of classifiers,
respectively for predicting activity class and completeness.
With this framework, we achieved significant performance gain over state-of-the-art
methods on both ActivityNet and THUMOS14.
Moreover, we demonstrated that our method is both accurate and generic,
being able to localize temporal boundaries precisely and working well
for activity classes with very different temporal structures.
\vspace{-15pt}
\paragraph{Acknowledgment.} This work is partially supported by the Big Data Collaboration Research grant from SenseTime Group (CUHK Agreement No. TS1610626), the General Research Fund (GRF) of Hong Kong (No. 14236516) and the Early Career Scheme (ECS) of Hong Kong (No. 24204215).

{\small
	\bibliographystyle{ieee}
	\bibliography{action_detection}

\begin{thebibliography}{10}\itemsep=-1pt

\bibitem{Supplement}
Supplementary materials.
\newblock \url{http://yjxiong.me/others/ssn/supp.pdf}.

\bibitem{andriluka2009pictorial}
M.~Andriluka, S.~Roth, and B.~Schiele.
\newblock Pictorial structures revisited: People detection and articulated pose
  estimation.
\newblock In {\em CVPR}, pages 1014--1021. IEEE, 2009.

\bibitem{de2016online}
R.~De~Geest, E.~Gavves, A.~Ghodrati, Z.~Li, C.~Snoek, and T.~Tuytelaars.
\newblock Online action detection.
\newblock In {\em ECCV}, pages 269--284. Springer, 2016.

\bibitem{Deng2009ImageNet}
J.~Deng, W.~Dong, R.~Socher, L.~Li, K.~Li, and F.~Li.
\newblock {ImageNet}: {A} large-scale hierarchical image database.
\newblock In {\em CVPR}, pages 248--255, 2009.

\bibitem{DonahueJ2015LRCN}
J.~Donahue, L.~Anne~Hendricks, S.~Guadarrama, M.~Rohrbach, S.~Venugopalan,
  K.~Saenko, and T.~Darrell.
\newblock Long-term recurrent convolutional networks for visual recognition and
  description.
\newblock In {\em CVPR}, pages 2625--2634, 2015.

\bibitem{Escorcia2016DAP}
V.~Escorcia, F.~Caba~Heilbron, J.~C. Niebles, and B.~Ghanem.
\newblock Daps: Deep action proposals for action understanding.
\newblock In B.~Leibe, J.~Matas, N.~Sebe, and M.~Welling, editors, {\em ECCV},
  pages 768--784, 2016.

\bibitem{caba2016cvpr}
B.~G. Fabian Caba~Heilbron, Juan Carlos~Niebles.
\newblock Fast temporal activity proposals for efficient detection of human
  actions in untrimmed videos.
\newblock In {\em CVPR}, pages 1914--1923, 2016.

\bibitem{caba2015activitynet}
B.~G. Fabian Caba~Heilbron, Victor~Escorcia and J.~C. Niebles.
\newblock Activitynet: A large-scale video benchmark for human activity
  understanding.
\newblock In {\em CVPR}, pages 961--970, 2015.

\bibitem{Felzenszwalb2010DPM}
P.~F. Felzenszwalb, R.~B. Girshick, D.~McAllester, and D.~Ramanan.
\newblock Object detection with discriminatively trained part-based models.
\newblock {\em IEEE TPAMI}, 32(9):1627--1645, 2010.

\bibitem{Fernando15Evolution}
B.~Fernando, E.~Gavves, J.~O. M., A.~Ghodrati, and T.~Tuytelaars.
\newblock Modeling video evolution for action recognition.
\newblock In {\em CVPR}, pages 5378--5387, 2015.

\bibitem{Gaidon2013Actom}
A.~Gaidon, Z.~Harchaoui, and C.~Schmid.
\newblock Temporal localization of actions with actoms.
\newblock {\em IEEE TPAMI}, 35(11):2782--2795, 2013.

\bibitem{Girshick2015FRCNN}
R.~Girshick.
\newblock Fast r-cnn.
\newblock In {\em ICCV}, pages 1440--1448, 2015.

\bibitem{Girshick2014RCNN}
R.~Girshick, J.~Donahue, T.~Darrell, and J.~Malik.
\newblock Rich feature hierarchies for accurate object detection and semantic
  segmentation.
\newblock In {\em CVPR}, pages 580--587, 2014.

\bibitem{Gkioxari2015Tubes}
G.~Gkioxari and J.~Malik.
\newblock Finding action tubes.
\newblock In {\em CVPR}, pages 759--768, June 2015.

\bibitem{He2014SPP}
K.~He, X.~Zhang, S.~Ren, and J.~Sun.
\newblock Spatial pyramid pooling in deep convolutional networks for visual
  recognition.
\newblock In {\em ECCV}, pages 346--361. Springer, 2014.

\bibitem{Hoai2011Joint}
M.~Hoai, Z.-Z. Lan, and F.~De~la Torre.
\newblock Joint segmentation and classification of human actions in video.
\newblock In {\em CVPR}, pages 3265--3272. IEEE, 2011.

\bibitem{hoiem2008putting}
D.~Hoiem, A.~A. Efros, and M.~Hebert.
\newblock Putting objects in perspective.
\newblock {\em IJCV}, 80(1):3--15, 2008.

\bibitem{JainCVPR14Tubelet}
M.~Jain, J.~C. van Gemert, H.~J\'egou, P.~Bouthemy, and C.~G.~M. Snoek.
\newblock Action localization by tubelets from motion.
\newblock In {\em CVPR}, June 2014.

\bibitem{Jiang2014THUMOS14}
Y.-G. Jiang, J.~Liu, A.~Roshan~Zamir, G.~Toderici, I.~Laptev, M.~Shah, and
  R.~Sukthankar.
\newblock {THUMOS} challenge: Action recognition with a large number of
  classes.
\newblock \url{http://crcv.ucf.edu/THUMOS14/}, 2014.

\bibitem{KarpathyCVPR14Sports1M}
A.~Karpathy, G.~Toderici, S.~Shetty, T.~Leung, R.~Sukthankar, and L.~Fei-Fei.
\newblock Large-scale video classification with convolutional neural networks.
\newblock In {\em CVPR}, pages 1725--1732, 2014.

\bibitem{lafferty2001conditional}
J.~Lafferty, A.~McCallum, F.~Pereira, et~al.
\newblock Conditional random fields: Probabilistic models for segmenting and
  labeling sequence data.
\newblock In {\em ICML}, volume~1, pages 282--289, 2001.

\bibitem{Laptev05STIP}
I.~Laptev.
\newblock On space-time interest points.
\newblock {\em IJCV}, 64(2-3), 2005.

\bibitem{Lazebnik2006Beyond}
S.~Lazebnik, C.~Schmid, and J.~Ponce.
\newblock Beyond bags of features: Spatial pyramid matching for recognizing
  natural scene categories.
\newblock In {\em CVPR}, volume~2, pages 2169--2178. IEEE, 2006.

\bibitem{Li2016RFCN}
Y.~Li, K.~He, J.~Sun, et~al.
\newblock R-fcn: Object detection via region-based fully convolutional
  networks.
\newblock In {\em NIPS}, pages 379--387, 2016.

\bibitem{Mettes2015Bofrag}
P.~Mettes, J.~C. van Gemert, S.~Cappallo, T.~Mensink, and C.~G. Snoek.
\newblock Bag-of-fragments: Selecting and encoding video fragments for event
  detection and recounting.
\newblock In {\em ICMR}, pages 427--434, 2015.

\bibitem{mettes2016spot}
P.~Mettes, J.~C. van Gemert, and C.~G. Snoek.
\newblock Spot on: Action localization from pointly-supervised proposals.
\newblock In {\em ECCV}, pages 437--453. Springer, 2016.

\bibitem{Montes_2016_NIPSWS}
A.~Montes, A.~Salvador, S.~Pascual, and X.~Giro-i Nieto.
\newblock Temporal activity detection in untrimmed videos with recurrent neural
  networks.
\newblock In {\em NIPS Workshop}, 2016.

\bibitem{Ng15BeyondSnippet}
J.~Y.-H. Ng, M.~Hausknecht, S.~Vijayanarasimhan, O.~Vinyals, R.~Monga, and
  G.~Toderici.
\newblock Beyond short snippets: Deep networks for video classification.
\newblock In {\em CVPR}, pages 4694--4702, 2015.

\bibitem{Niebles2010Modeling}
J.~C. Niebles, C.-W. Chen, and L.~Fei-Fei.
\newblock Modeling temporal structure of decomposable motion segments for
  activity classification.
\newblock In {\em ECCV}, pages 392--405. Springer, 2010.

\bibitem{Oneata2013FV}
D.~Oneata, J.~Verbeek, and C.~Schmid.
\newblock Action and event recognition with fisher vectors on a compact feature
  set.
\newblock In {\em ICCV}, pages 1817--1824, 2013.

\bibitem{oneata2014lear}
D.~Oneata, J.~Verbeek, and C.~Schmid.
\newblock The lear submission at thumos 2014.
\newblock In {\em THUMOS Action Recognition Challenge}, 2014.

\bibitem{Peng2016ActionFRCNN}
X.~Peng and C.~Schmid.
\newblock Multi-region two-stream r-cnn for action detection.
\newblock In {\em ECCV}. Springer, 2016.

\bibitem{Pirsiavash2014SegGrammar}
H.~Pirsiavash and D.~Ramanan.
\newblock Parsing videos of actions with segmental grammars.
\newblock In {\em CVPR}, pages 612--619, 2014.

\bibitem{PT2015MCG}
J.~Pont-Tuset, P.~Arbel\'{a}ez, J.~Barron, F.~Marques, and J.~Malik.
\newblock Multiscale combinatorial grouping for image segmentation and object
  proposal generation.
\newblock In {\em arXiv:1503.00848}, March 2015.

\bibitem{Ren2015FasterRCNN}
S.~Ren, K.~He, R.~Girshick, and J.~Sun.
\newblock Faster r-cnn: Towards real-time object detection with region proposal
  networks.
\newblock In {\em NIPS}, pages 91--99, 2015.

\bibitem{Richard2016Language}
A.~Richard and J.~Gall.
\newblock Temporal action detection using a statistical language model.
\newblock In {\em CVPR}, pages 3131--3140, 2016.

\bibitem{Roerdink2000Watershed}
J.~B. Roerdink and A.~Meijster.
\newblock The watershed transform: Definitions, algorithms and parallelization
  strategies.
\newblock {\em Fundamenta informaticae}, 41(1, 2):187--228, 2000.

\bibitem{Shou2016SCNN}
Z.~Shou, D.~Wang, and S.-F. Chang.
\newblock Temporal action localization in untrimmed videos via multi-stage
  {CNNs}.
\newblock In {\em CVPR}, pages 1049--1058, 2016.

\bibitem{Shrivastava2016OHNM}
A.~Shrivastava, A.~Gupta, and R.~Girshick.
\newblock Training region-based object detectors with online hard example
  mining.
\newblock In {\em CVPR}, pages 761--769, 2016.

\bibitem{Simonyan14TwoStream}
K.~Simonyan and A.~Zisserman.
\newblock Two-stream convolutional networks for action recognition in videos.
\newblock In {\em NIPS}, pages 568--576, 2014.

\bibitem{Singh_2016_CVPR}
B.~Singh, T.~K. Marks, M.~Jones, O.~Tuzel, and M.~Shao.
\newblock A multi-stream bi-directional recurrent neural network for
  fine-grained action detection.
\newblock In {\em CVPR}, pages 1961--1970, 2016.

\bibitem{DBLP:journals/corr/SinghC16}
G.~Singh and F.~Cuzzolin.
\newblock Untrimmed video classification for activity detection: submission to
  activitynet challenge.
\newblock {\em CoRR}, abs/1607.01979, 2016.

\bibitem{Soomro2012Ucf101}
K.~Soomro, A.~R. Zamir, and M.~Shah.
\newblock Ucf101: A dataset of 101 human actions classes from videos in the
  wild.
\newblock {\em arXiv:1212.0402}, 2012.

\bibitem{Tang2013RightFeature}
K.~Tang, B.~Yao, L.~Fei-Fei, and D.~Koller.
\newblock Combining the right features for complex event recognition.
\newblock In {\em CVPR}, pages 2696--2703, 2013.

\bibitem{Tran15C3D}
D.~Tran, L.~D. Bourdev, R.~Fergus, L.~Torresani, and M.~Paluri.
\newblock Learning spatiotemporal features with {3D} convolutional networks.
\newblock In {\em ICCV}, pages 4489--4497, 2015.

\bibitem{Van2011SS}
K.~E. Van~de Sande, J.~R. Uijlings, T.~Gevers, and A.~W. Smeulders.
\newblock Segmentation as selective search for object recognition.
\newblock In {\em ICCV}, pages 1879--1886, 2011.

\bibitem{WangS13IDT}
H.~Wang and C.~Schmid.
\newblock Action recognition with improved trajectories.
\newblock In {\em ICCV}, pages 3551--3558, 2013.

\bibitem{wang2014action}
L.~Wang, Y.~Qiao, and X.~Tang.
\newblock Action recognition and detection by combining motion and appearance
  features.
\newblock In {\em THUMOS Action Recognition Challenge}, 2014.

\bibitem{Wang2014Latent}
L.~Wang, Y.~Qiao, and X.~Tang.
\newblock Latent hierarchical model of temporal structure for complex activity
  classification.
\newblock {\em IEEE TIP}, 23(2):810--822, 2014.

\bibitem{WangQT15TDD}
L.~Wang, Y.~Qiao, and X.~Tang.
\newblock Action recognition with trajectory-pooled deep-convolutional
  descriptors.
\newblock In {\em CVPR}, pages 4305--4314, 2015.

\bibitem{Wang2016Actionness}
L.~Wang, Y.~Qiao, X.~Tang, and L.~Van~Gool.
\newblock Actionness estimation using hybrid fully convolutional networks.
\newblock In {\em CVPR}, pages 2708--2717, 2016.

\bibitem{Wang2016TSN}
L.~Wang, Y.~Xiong, Z.~Wang, Y.~Qiao, D.~Lin, X.~Tang, and L.~Van~Gool.
\newblock Temporal segment networks: Towards good practices for deep action
  recognition.
\newblock In {\em ECCV}, pages 20--36, 2016.

\bibitem{Wang2016TPP}
P.~Wang, Y.~Cao, C.~Shen, L.~Liu, and H.~T. Shen.
\newblock Temporal pyramid pooling based convolutional neural network for
  action recognition.
\newblock {\em IEEE TCSVT}, 2016.

\bibitem{UTS}
R.~Wang and D.~Tao.
\newblock {UTS} at activitynet 2016.
\newblock In {\em AcitivityNet Large Scale Activity Recognition Challenge
  2016}, 2016.

\bibitem{Weinzaepfel2015Track}
P.~Weinzaepfel, Z.~Harchaoui, and C.~Schmid.
\newblock Learning to track for spatio-temporal action localization.
\newblock In {\em ICCV}, pages 3164--3172, 2015.

\bibitem{Yeung2016FrameGlimpse}
S.~Yeung, O.~Russakovsky, G.~Mori, and L.~Fei-Fei.
\newblock End-to-end learning of action detection from frame glimpses in
  videos.
\newblock In {\em CVPR}, pages 2678--2687, 2016.

\bibitem{Yuan2016ScorePyramids}
J.~Yuan, B.~Ni, X.~Yang, and A.~A. Kassim.
\newblock Temporal action localization with pyramid of score distribution
  features.
\newblock In {\em CVPR}, pages 3093--3102, 2016.

\bibitem{ZhangWW0W16}
B.~Zhang, L.~Wang, Z.~Wang, Y.~Qiao, and H.~Wang.
\newblock Real-time action recognition with enhanced motion vector {CNNs}.
\newblock In {\em CVPR}, pages 2718--2726, 2016.

\bibitem{Dollar2014Edgebox}
C.~L. Zitnick and P.~Doll{\'{a}}r.
\newblock Edge boxes: Locating object proposals from edges.
\newblock In {\em ECCV}, pages 391--405, 2014.

\end{thebibliography}
}

\end{document}